%% file: paper.tex
\newcommand{\CLASSINPUTbaselinestretch}{1}
\documentclass[conference]{IEEEtran}
\usepackage{etoolbox}
\makeatletter
\patchcmd{\maketitle}{\@copyrightspace}{}{}{}

\makeatother
\usepackage{xspace}
\usepackage{siunitx}
\usepackage[]{graphicx}
\usepackage{stfloats}  
\usepackage{cite}  
\usepackage{psfrag}  
\usepackage{subfigure}
\usepackage{wrapfig}
\usepackage{epsfig}
\usepackage{url}
\usepackage{amsmath, amssymb}
\usepackage{array}
\usepackage{soul}
\usepackage{multicol}
\usepackage{algorithm}
\usepackage{algorithmic}
\usepackage{mathrsfs}
\usepackage{multirow}
\usepackage[normalem]{ulem}
\usepackage{verbatim}
\usepackage{fixltx2e}
\usepackage[utf8x]{inputenc} 
\usepackage{ucs} 
\usepackage{amsfonts} 
\usepackage{makeidx} 
\usepackage[dvipsnames,usenames]{color} 
\usepackage{array} 
\usepackage{colortbl} 
\usepackage{booktabs} 
\usepackage{amsmath}
\usepackage{comment}
\usepackage[justification=centering]{caption}

\usepackage{pifont}% http://ctan.org/pkg/pifont
\newcommand{\name}{PIM-DRAM: Accelerating Machine Learning Workloads using Processing in Commodity DRAM}%
\definecolor{tableheading}{rgb}{0.9,0.9,0.9}
\definecolor{softblue}{rgb}{0.8,0.8,1} 
\arrayrulecolor{Black} 
\usepackage{tabu}
\usepackage{floatflt}
\usepackage{amsmath, amssymb}
\usepackage{tikz}
\usepackage{multicol}
\usepackage{amsmath,amssymb}
\usepackage{tabularx}
\makeatletter
\newcommand*{\textoverline}[1]{$\overline{\hbox{#1}}\m@th$}
\makeatother

\begin{document}

\title{
  \name}
\author{\IEEEauthorblockN{Sourjya Roy, Mustafa Ali and Anand Raghunathan}
School of Electrical and Computer Engineering, Purdue University, West Lafayette, IN, USA\\ }
\begingroup\renewcommand\thefootnote{\textsection}
%\footnotetext{\textsuperscript{*}}
%\footnotetext{{\textdagger }}
\endgroup

\sloppy

\maketitle

\input{sections/abstract}

%\keywords{}
%%%%%%%%%%%%%%%%%%%%%%%%%%%%%%%%%%%%%%%%%%%%%%%%%%%%%%%%%%%% Section
%\vspace*{-10pt}
\section{Introduction}
\label{sec:introduction}
\input{sections/Introduction}
\vspace*{-0pt}

%%%%%%%%%%%%%%%%%%%%%%%%%%%%%%%%%%%%%%%%%%%%%%%%%%%%%%%%%%%% Section

\vspace*{6pt}
\section{Background}
\label{sec:preliminaries}
\input{sections/Preliminaries.tex}
\vspace*{-0pt}

%%%%%%%%%%%%%%%%%%%%%%%%%%%%%%%%%%%%%%%%%%%%%%%%%%%%%%%%%%%% Section
\vspace*{4pt}
\section{Proposed in-DRAM Computing Primitives}
\label{sec:tool}
\input{sections/tool}

%%%%%%%%%%%%%%%%%%%%%%%%%%%%%%%%%%%%%%%%%%%%%%%%%%%%%%%%%%%% Section
\vspace*{2pt}
\section{Architecture and Data Flow}

\label{sec:exptsetup}
\input{sections/exptsetup}

%%%%%%%%%%%%%%%%%%%%%%%%%%%%%%%%%%%%%%%%%%%%%%%%%%%%%%%%%%%% Section
\vspace*{10pt}
\section{Evaluation Methodology and Results}
\label{sec:results}
\input{sections/results.tex}

%%%%%%%%%%%%%%%%%%%%%%%%%%%%%%%%%%%%%%%%%%%%%%%%%%%%%%%%%%%% Section
% \vspace*{2pt}
% \section{Related Work }
% \label{sec:relatedWork}
% \input{sections/relatedWork}

%%%%%%%%%%%%%%%%%%%%%%%%%%%%%%%%%%%%%%%%%%%%%%%%%%%%%%%%%%%% Section

\vspace*{6pt}
\section{Conclusion}
\label{sec:conclusion}
\input{sections/conclusion}

%%%%%%%%%%%%%%%%%%%%%%%%%%%%%%%%%%%%%%%%%%%%%%%%%%%%%%%%%%%% Section
\section{Acknowledgment}
This work was supported by C-BRIC, one of six centers in JUMP, a Semiconductor Research Corporation (SRC) program, sponsored by DARPA.
\vspace*{-0pt}
\scriptsize
\bibliographystyle{IEEEtran}
\bibliography{paper}

%--\normalsize
%\newpage
%\appendix
%%\input{suppsec}
%\input{biographies}

\end{document}

%% file: sections/abstract.tex
\begin{abstract}
Deep Neural Networks (DNNs) have transformed the field of machine learning and are widely deployed in many applications involving image, video, speech and natural language processing. The increasing compute demands of DNNs have been widely addressed through Graphics Processing Units (GPUs) and specialized accelerators. However, as model sizes grow, these von Neumann architectures require very high memory bandwidth to keep the processing elements utilized as a majority of the data resides in the main memory. Processing in memory has been proposed as a promising solution for the memory wall bottleneck for ML workloads. In this work, we propose a new DRAM-based processing-in-memory (PIM) multiplication primitive coupled with intra-bank accumulation to accelerate matrix vector operations in ML workloads. The proposed multiplication primitive adds $<$1\% area overhead and does not require any change in the DRAM peripherals. Therefore, the proposed multiplication can be easily adopted in commodity DRAM chips. Subsequently, we design a DRAM-based PIM architecture, data mapping scheme and dataflow for executing DNNs within DRAM. System evaluations performed on networks like AlexNet, VGG16 and ResNet18 show that the proposed architecture, mapping, and data flow can provide up to 19.5x speedup over an NVIDIA Titan Xp GPU highlighting the need to overcome the memory bottleneck in future generations of DNN hardware.

%and 6.5x benefits over a GPU and an ideal conventional (non-PIM) baseline architecture with infinite compute bandwidth, respectively.
%% what is non-PIM?? GPU is also non-PIM. non-PIM-DRAM? Not clear at all.
%% what kind of GPU? with standard DRAM?
\end{abstract}

% 150 word limit for IEEE Micro

%% file: sections/Introduction.tex
{\noindent}
Machine learning has extensively proliferated into several application domains including visual processing, speech recognition, health care and autonomous driving \cite{fortune-dnns,speech}. Modern ML workloads perform billions of computations and require an ever-increasing amount of memory to store inputs, weights and outputs. As a result, domain-specific hardware such as Graphics Processing Units (GPUs) and Tensor Processing Units (TPUs) have emerged as state-of-the-art ML accelerators \cite{tpu}. These accelerators provide substantially higher performance than CPUs. However, such accelerators can only reach their peak performance when they can fetch data from the main memory as fast as their compute throughput. This requires the Dynamic Random Access Memory (DRAM) to have both high bandwidth and high capacity. However, current DRAM technologies are struggling to keep up with the bandwidth and capacity requirements of modern ML workloads \cite{scaledeep}.
\begin{figure}[htb]
    \centering
    \vspace*{-2pt} 
    \includegraphics[width=0.8\columnwidth]{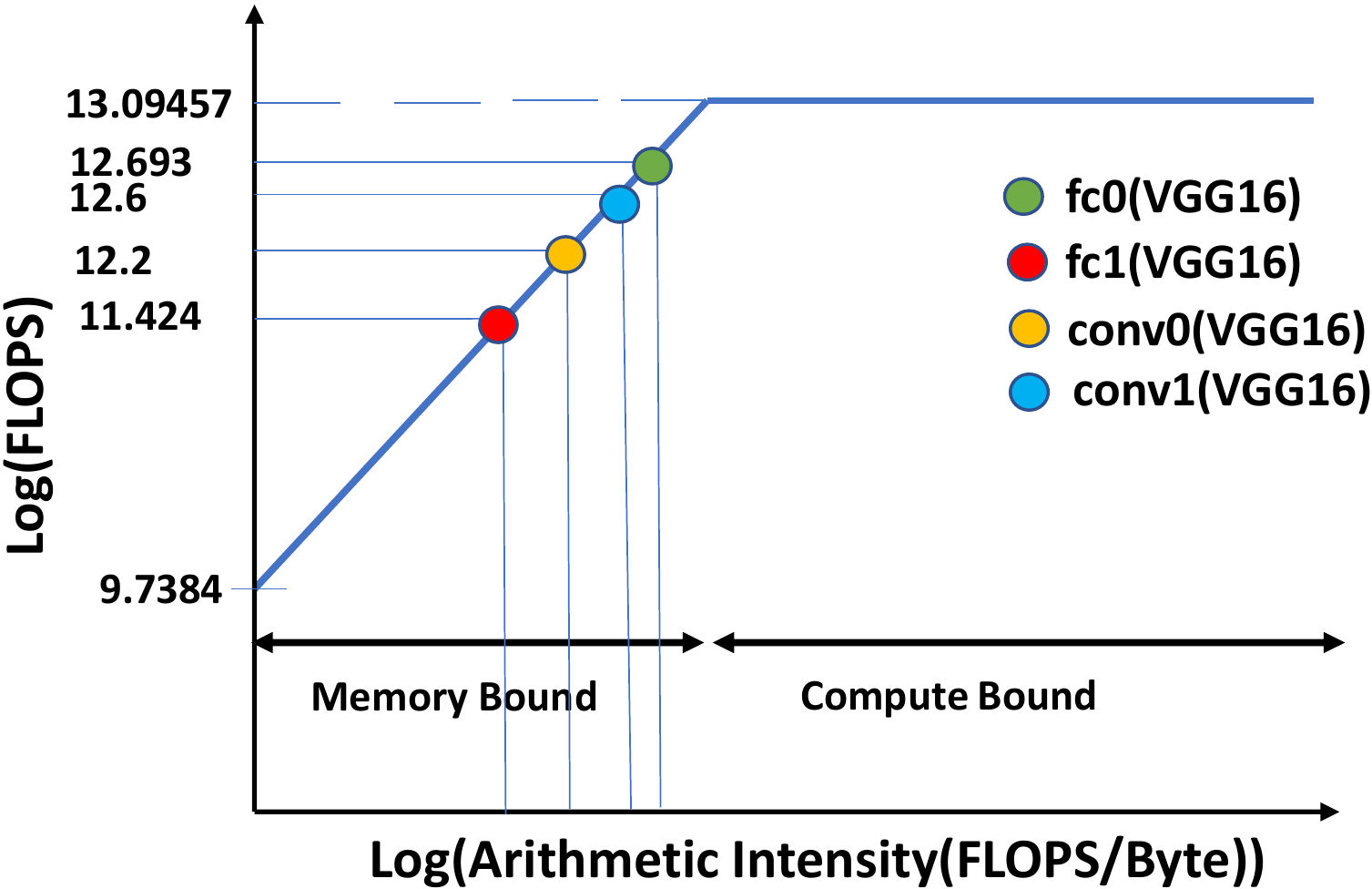}
    \caption{TITAN Xp roofline for VGG16}
	\vspace*{-2pt}
    \label{fig:roofline}
\end{figure}

% --The need for PIM and PIM categories (PIM, PNM, …)

Fig \ref{fig:roofline} shows that the some of layers of a popular network like VGG16 lies in the memory bound part of the roofline model for NVIDIA TITAN Xp GPU. Hence, Processing in Memory (PIM) based on DRAM can be a promising solution to reduce the off-chip memory bandwidth requirements of ML workloads. A normal DRAM access fetches data to/from a single bank due to the limited bandwidth of the memory channel between the host and the memory. However, PIM achieves higher memory bandwidth since data can be accessed and computed in parallel in all subarrays/banks through the significantly-wider internal DRAM buses. Therefore, by processing in DRAM, the amount of data transferred between the host side and memory is amortized. PIMs can be categorized into three main variants based on where the computation is performed: (1) In-subarray computing, where the compute occurs at the local sense amplifiers \cite{mustafa,simdram,elp2im,drisa,dracc}, (2) Near-bank computing, where compute blocks are added just after the sense amplifiers \cite{simdram,newton}, and (3) 3D-based PIM, where compute cores are put on the logic die in a Hybrid Memory Cube (HMC)-like architecture \cite{hmc,neurocube,tetris}.

In-subarray computing exposes the maximum internal DRAM bandwidth since the compute operation occurs at the local sense amplifiers leading to higher performance benefits over near-bank and 3D PIM variants. The processing in subarray happens by enabling multiple wordlines at the same time \cite{ambit,drisa,dracc, elp2im,mustafa} to leverage the large available internal DRAM bandwidth. However, there are several challenges preventing wide adoption of existing in-subarray computing proposals for accelerating ML workloads. First, most of the previously-reported subarray PIM designs are only capable of simple bit-wise operations \cite{mustafa,ambit,elp2im} or arithmetic addition \cite{mustafa, dracc}. \cite{dracc} only implements a ternary weight network using DRAM bit-wise operations. However, ML workloads require multiplication and accumulation operations in addition to non-linear activation functions, which are not supported in prior PIM works. Second, accelerating ML workloads in DRAM requires careful data orchestration to minimize internal data movements between banks. Therefore, optimal dataflow, especially tailored for DRAM-based PIM, is needed to maximize the performance benefits of PIM. 

To that end, we propose PIM-DRAM: a new processing-in-DRAM architecture for end-to-end machine learning acceleration. Differing from previous efforts, our proposal performs multiplication operations at the subarray level, while the accumulation and activation functions are performed in our proposed PIM-DRAM bank architecture. Moreover, we present specifically-tailored data orchestration of PIM-DRAM to maximize its benefits.
%The proposed in-subarray multiplication comprises bit-wise AND operation followed by column-based addition.
Our main contributions are as follows:
%% In the above, write a sentence on your PIM and suitability for ML workloads

\begin{itemize}
    \item We propose a novel multiplication scheme inside DRAM at the subarray level with negligible changes to the DRAM subarrays. Since the multiplication is performed by addition and AND operations, we propose a fast light-weight bit-wise in-subarray AND operation in DRAM to reduce the overall multiplication cost. Note, the proposed multiplication primitive adds only $<$1\% area overhead to a conventional DRAM chip and requires no modification to the DRAM peripheral circuitry.
    %% what is AND scheme?

    \item We incorporate novel PIM-DRAM bank architecture including adder trees and non-linear activation function units in each DRAM bank for efficient ML acceleration. The proposed architecture supports ReLU, batch-normalization, and pooling operations needed for a wide range of ML models.
    
    \item We introduce a specific mapping and dataflow for our proposed DRAM PIM to achieve high utilization and system-level performance.
    
    \item We perform extensive analyses of the proposed DRAM PIM architecture at both the circuit and architecture levels and compare it with a GPU.
    %% not clear what a non-PIM system is!!
\end{itemize}

% ----Data mapping for implementing whole networks

% --state the contributions

% 1. ----Contribution1 (AND/MUL primitive)

% 2. ----Contribution2 (near bank compute)

% 3. ----Contribution3 (workload mapping

%% file: sections/Preliminaries.tex
\noindent  
{\bf\noindent } 
This section provides background on conventional DRAM organization, basic DRAM read and write operations and other concepts pertinent to understanding the proposed DRAM-based PIM.
\subsection{Conventional DRAM Organization}
\vspace*{0pt}
 \begin{figure}[htb]
    \centering
    \vspace*{-2pt} 
    \includegraphics[width=0.8\columnwidth]{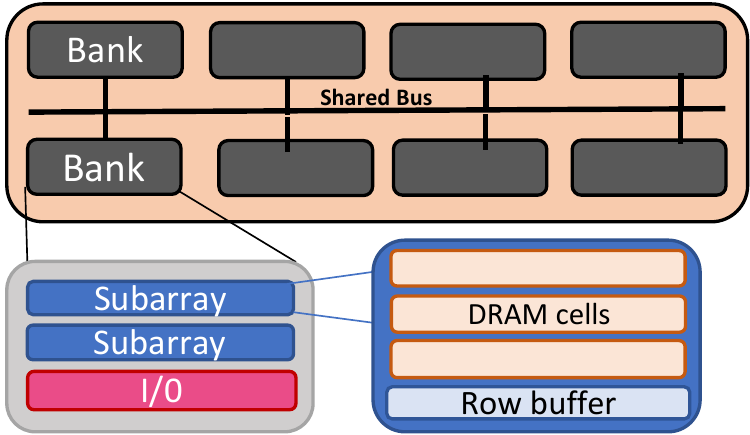}
    \caption{DRAM Hierarchy}
	\vspace*{-2pt}
    \label{fig:dramheir}
\end{figure}

The DRAM hierarchy consists of channels, modules and ranks. Further, a rank consists of multiple banks. Each bank consists of several subarrays. An example DRAM organization is shown in Fig \ref{fig:dramheir}.
\vspace*{0pt}
\begin{figure}[htb]
    \centering
    \vspace*{-2pt} 
    \includegraphics[width=0.8\columnwidth]{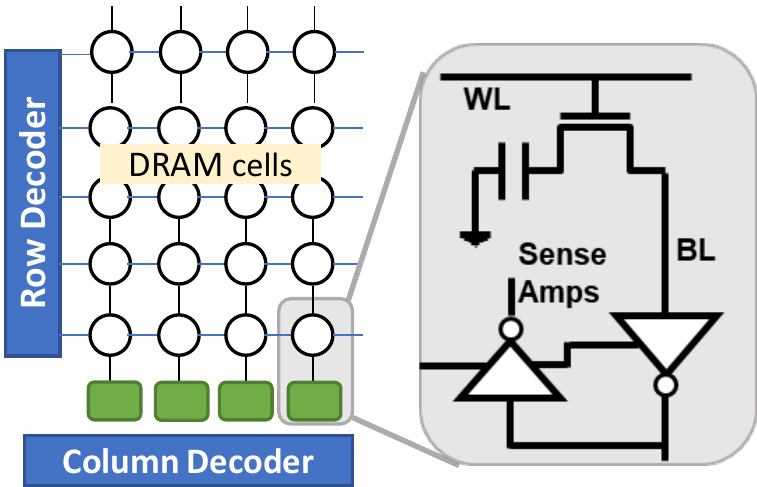}
    \caption{DRAM subarray view}
	\vspace*{-2pt}
    \label{fig:dramsub}
\end{figure}
Now, let us consider the internal organization of a DRAM bank as shown in Fig \ref{fig:dramsub}. One DRAM bank comprises of multiple subarrays, a subarray being a 2D array of DRAM cells with associated peripheral circuits. A DRAM cell consists of a storage capacitor with an access transistor connecting it to the bitline (BL) in a manner that is controlled by the wordline (WL) \cite{ambit}.
%% Provide a reference. Can be Rabaey's book.
The DRAM cells in the same row share a wordline and cells in the same column share a bitline. A voltage of magnitude VDD across the storage capacitor signifies a value of 1. Similarly, a magnitude of 0 V across the capacitor signifies a value of 0. To perform a DRAM read operation, BLs are initially precharged to VDD/2 by the PRECHARGE command. The address of the DRAM row to be read is applied to the row decoder that activates the corresponding WL. Once the correct WL is activated, charge sharing occurs between the bitline capacitance and the DRAM cell capacitor. The BL voltage is amplified by turning on the sense amplifiers that regenerate the BL voltage to either 0V or VDD when the cell data is "0" or "1", respectively. 
%% Not sure exactly what is meant by the above sentence.
The ACTIVATE command from the memory controller leads to activating the appropriate WL followed by enabling the sense amplifiers to read the row data.
%% Not clear about the above sentence
The column decoder selects the word bits to be read from the Sense Amplifiers. The write operation is similar to the read operation, where the column decoder now passes the word to be written to the corresponding BLs. Subsequently, the row decoder activates the corresponding WL for the row that is to be written.
\subsection{DRAM In-Memory Primitives}
Previous works have proposed in-subarray computing primitives supporting row copy \cite{rowclone}, bit-wise logic operations \cite{ambit} and bit-serial arithmetic addition \cite{mustafa}. 

RowClone \cite{rowclone} supports copying of data from a source row to a destination row intra-subarray, intra-bank, and inter-bank. Intra-subarray copies are achieved by either performing a complex subarray operation wherein we enable the source row WL, activate the sense amplifiers, and also enable the destination row. Alternatively, we leverage the existing interconnects across subarrays and across banks. We adopt the RowClone approach in this work to transfer data among banks.

Next, we describe the in-memory primitive for ADD in DRAM cells used for this work \cite{mustafa}. The ADD operation is based on the following 2 equations that leverage the multiple-row activation principle to compute the majority function. 
    \begin{equation}
        Cout=Majority(A,B,Cin)
    \end{equation}
    \begin{equation}
        Sum=Majority(A,B,Cin,\overline{Cout},\overline{Cout})
    \end{equation}
    
\begin{figure}[t!]
    \centering
    \includegraphics[width=\columnwidth]{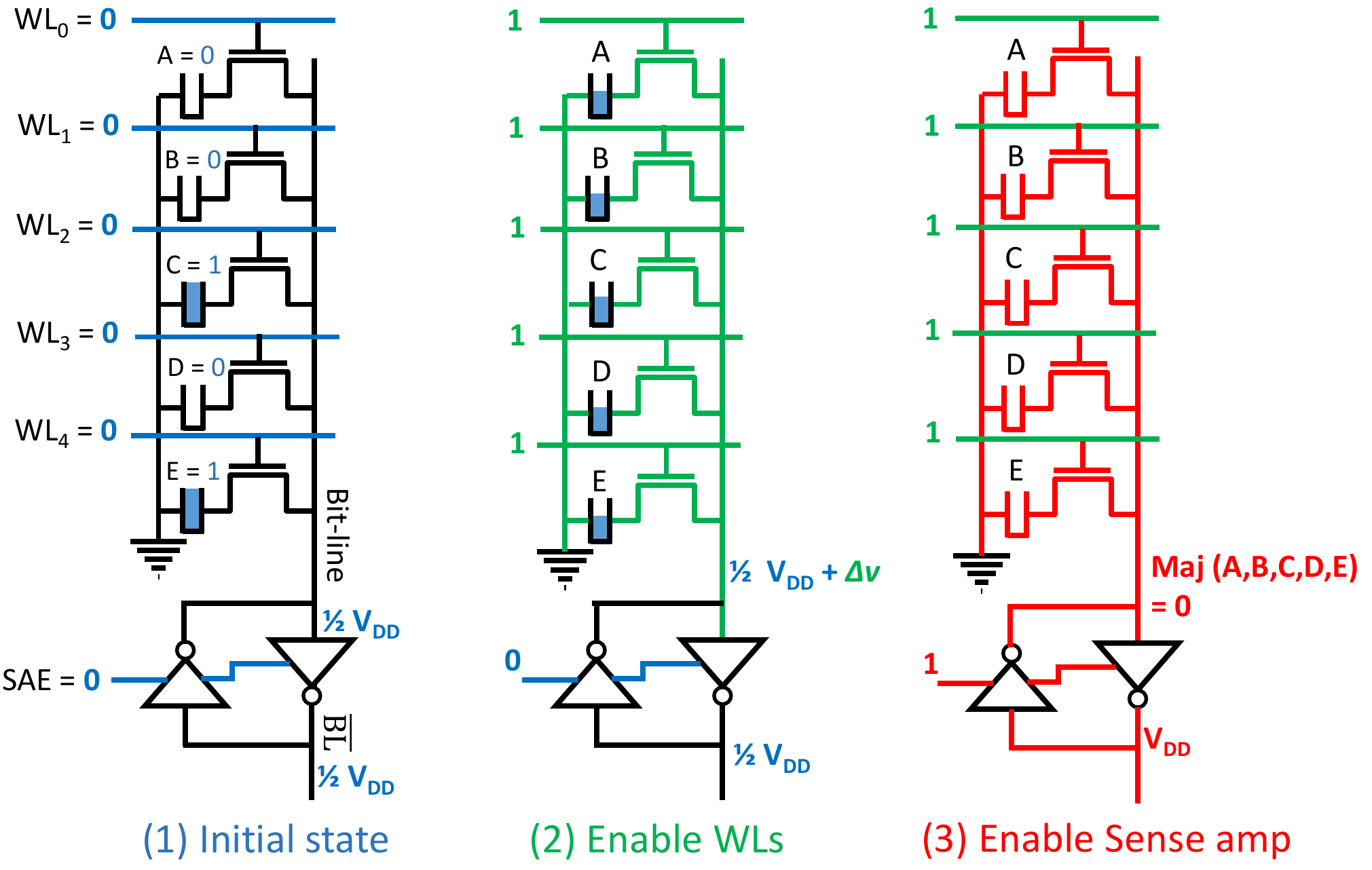}
    \caption{Quintuple-row activation: activating five rows at the same time to calculate the five-input majority function, proposed in \cite{mustafa}}.
    \label{fig:quintuple-activation}
\end{figure}

The ADD operation requires 9 additional "compute rows" in each sub-array for copying the original data and storing the carry-out and carry-in values. The 9 compute rows are used in the proposed Multiplication primitive as well, which is described in the later section. The operands A, B and carry-in (Cin) are copied to the compute rows. The addition operation comprises four main steps: (i) copy the first vector bit (A) to the compute rows. (ii) copy the second vector bit (B) to the compute rows. (iii) calculate Cout using the multiple-row activation scheme as shown in equation (1). (iv) calculate Sum using the multiple-row activation as shown in equation (2) and Fig. \ref{fig:quintuple-activation}. Note that, addition of two n-bit operands requires 4n+1 ACTIVATE-ACTIVATE-PRECHARGE (AAP) operations.
%% Show a figure for ADD -- similar to a slide that I have from Mustafa.

\subsection{DNN Workloads}
    \begin{figure}[htb]
    \centering
    \vspace*{-2pt} 
    \includegraphics[width=0.9\columnwidth]{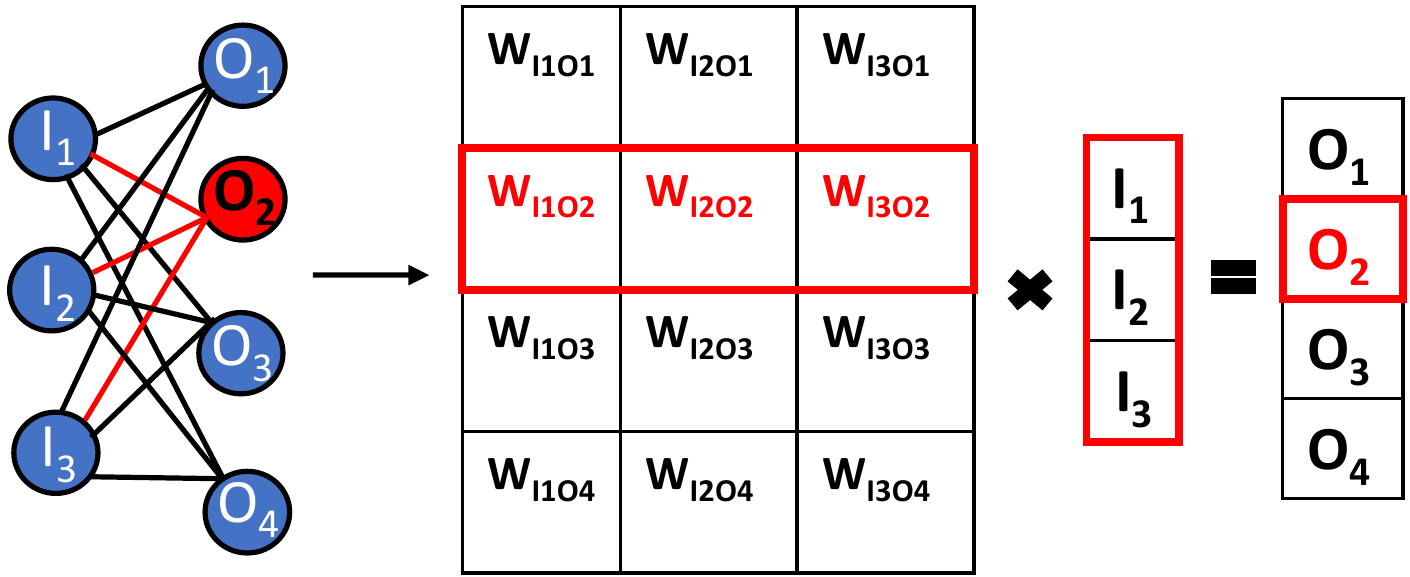}
    \caption{DNN Workload Example}
	\vspace*{-2pt}
    \label{fig:dnnworkload}
\end{figure}

Deep Neural Networks (DNNs) are adopted in a wide range of applications and are comprised of various primitives like Fully Connected (FC) Layers, Convolutional (CONV) Layers, and/or with recurrent connections as in Recurrent Neural Networks (RNNs). The dominant computation common across all the different kinds of network layers is Matrix Vector Multiplication (MVM), which is in turn comprised of Multiply and Accumulate operations (MACs). An example FC layer is shown in Fig \ref{fig:dnnworkload}.
%% Since you are hard pressed for space, do you need this figure?
The figure shows 3 input neurons being connected to 4 output neurons. Each of the output neurons have weight connections associated with each input. The input values are multiplied with their corresponding weight values and added together to calculate the final output value. The computation of the example here can be realized as 4 dot-products to compute the 4 output neuron values, each of which requires 3 Multiply-Accumulate operations.

%% file: sections/tool.tex
{\noindent} 

To perform multiplication in DRAM, we perform bit-wise AND operations between the operands bits, followed by addition of the AND outputs. This section describes the proposed in-DRAM AND followed by the overall multiplication computation in DRAM subarrays.
\subsection{Bit-Wise AND Operation}

\begin{comment}

\begin{figure}
    \includegraphics[width=.24\textwidth]{img/0,0.png}\hfill
    \includegraphics[width=.24\textwidth]{img/0,1.2.png}\hfill
    \\[\smallskipamount]
     \includegraphics[width=.24\textwidth]{img/1.2,0.png}\hfill
      \includegraphics[width=.24\textwidth]{img/1.2,0.png}
      \caption{This figure shows the simulation results for AND operation for (0,0) (top row left), (0,VDD) (top row right), (VDD,0) (bottom row left), (VDD,VDD) (bottom row right)}
      \label{fig:hspice}
\end{figure}
\end{comment}

\begin{figure}[htb]
    \centering
    \vspace*{-2pt} 
    \includegraphics[width=\columnwidth]{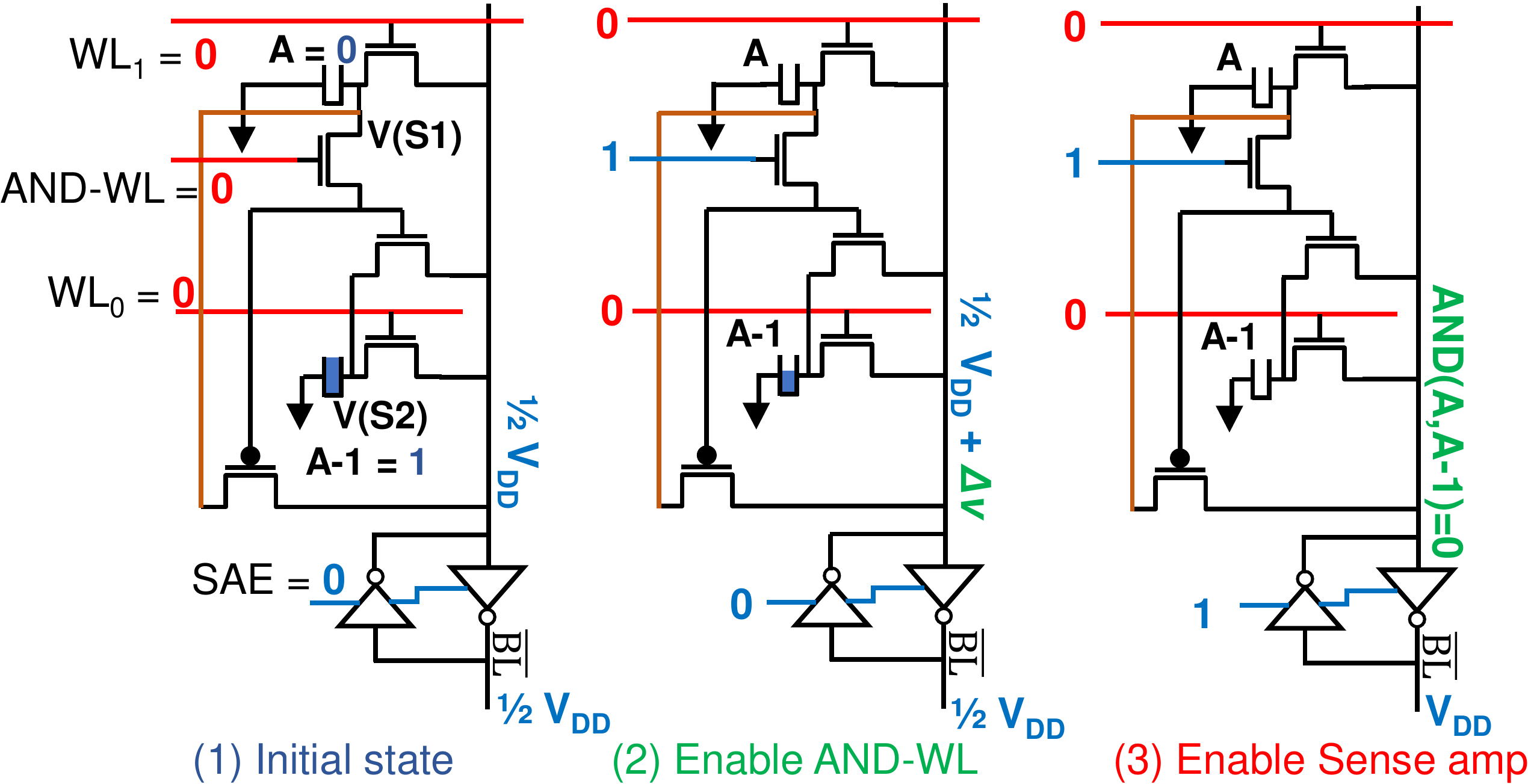}
    \caption{Example of AND operation}
	\vspace*{-2pt}
    \label{fig:and1}
\end{figure}
We perform the AND operation in a DRAM subarray as shown in Fig \ref{fig:and1}. We dedicate two extra rows, called compute rows (denoted by A and A-1 in the figure), into which we first copy the operands so as to maintain the original data. Initially, the operands are copied to A and A-1 using Rowclone \cite{rowclone}. The bitline is then precharged to VDD/2 and AND-WL is activated. Based on the data stored in A, cell capacitor A or A-1 gets connected to the BL through the PMOS and NMOS, respectively.
%%Anand: I don't see any PMOS transistor in the figure!
%% Charge sharing between what?
Subsequently, the sense amplifiers are turned on and the BL voltage gets amplified to either 0 or VDD based on the AND result.
%% the sense amplifier amplifies the change in the voltage. The charge does not get amplified.
Fig \ref{fig:and1} shows the three stages of the AND operation for the copied operands in rows A and A-1. It can be seen that each of these three stages is realized using an AAP operation. Three extra transistors is equivalent to three extra rows. The addition of nine compute rows with three extra transistors leads to very small ($<$ 1\%) area overhead at the subarray level.
%% AAP operations?
%% I did not read the next portion. Please read and update.

\subsection{In-DRAM Multiplication}
\begin{figure}[htb]
    \centering
    \vspace*{-2pt} 
    \includegraphics[width=0.65\columnwidth]{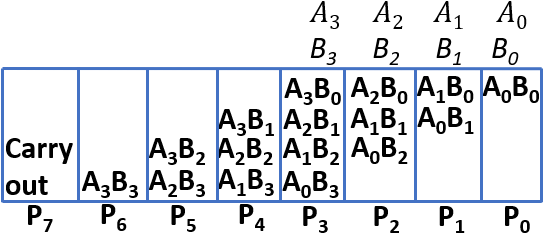}
    \caption{Multiplication operation}
	\vspace*{-2pt}
    \label{fig:mulex}
\end{figure}
The Multiplication operation can be broken down into AND and add operations. An example 4 bit multiplication is shown in Fig \ref{fig:mulex}. Here A\textsubscript{n} and B\textsubscript{n} refer to the n\textsuperscript{th} bit of the two operands and P\textsubscript{n} refers to the n\textsuperscript{th} bit of the product. It is evident from the figure that the AND results from every column needs to be added along with the carry in to get the P\textsubscript{n} and the carry out. As an example, A\textsubscript{1}B\textsubscript{0}, A\textsubscript{0}B\textsubscript{1} and the carry out of P\textsubscript{0} are added to get the result for P\textsubscript{1} and the carry required for P\textsubscript{2} computation.

\begin{figure*}[htb]
  \vspace*{-0pt}
  \centering
  \includegraphics[width=\textwidth]{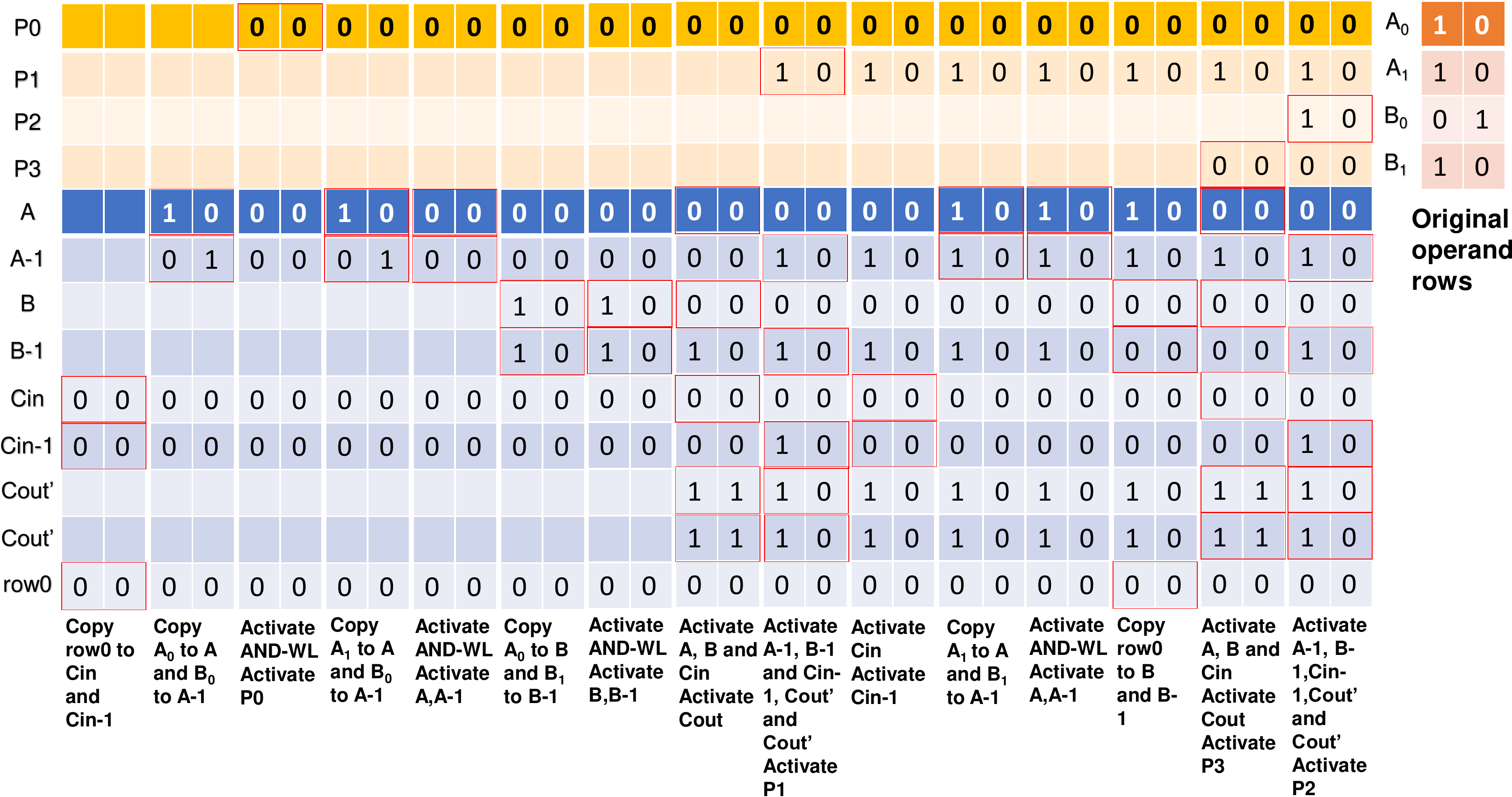}
  \vspace*{-2pt}
  \caption{Example of 2 bit Multiplication}
  \label{fig:MACeg}
  \vspace*{-8pt}
\end{figure*}

Next, we discuss realizing the multiplication operation in a DRAM subarray. The multiplication primitive requires 9 compute rows- A, A-1, B, B-1, Cin, Cin-1, Cout, Cout-1 and row0 as shown in Fig \ref{fig:MACeg}. These reserved compute rows account for $<$1\% of the total subarray size. A split-line decoder is used to activate the 9 compute rows. It is worth mentioning that the multiplication does not require any peripheral circuit modification, therefore, it can be easily adopted in commodity DRAMs.

We illustrate the proposed in-DRAM multiplication with the example of a 2 bit multiplication in Fig \ref{fig:MACeg}. row0 is used for storing 0's. At first row0 is copied to Cin and Cin-1. This serves as the carry in for the computing the LSB of the product. Next, the operands A\textsubscript{0} and B\textsubscript{0} are copied to A and A-1. Once the operands are copied, the AND-WL is activated followed by turning on the Sense Amplifiers. After Sense Amplification, P0 is activated to store the result of the AND operation. Next, A\textsubscript{1} and B\textsubscript{0} are copied to A and A-1 respectively. As before, the AND-WL is activated. After charge sharing and sense amplification, A and A-1 are activated and both store the result of A\textsubscript{1} AND B\textsubscript{0}. Next, A\textsubscript{0} and B\textsubscript{1} are copied to B and B-1 followed by activating the AND-WL. After charge sharing, the Sense Amplifiers are turned on and B and B-1 are activated. B and B-1 now store the result of A\textsubscript{0} AND  B\textsubscript{1}. After performing the two AND operations, the results of A\textsubscript{1}B\textsubscript{0} and A\textsubscript{0}B\textsubscript{1} need to be added. This primitive uses the majority based adder discussed in \cite{mustafa}. As we know from previous discussions, A\textsubscript{1}B\textsubscript{0} is present in A, A-1 and A\textsubscript{0}B\textsubscript{1} is present in B, B-1 because of the AND operation. Hence we do not need to copy the operands to the compute rows and this leads to fewer AAP operations than the add in \cite{mustafa}. For addition, triple row activation of A, B and Cin is performed to obtain the carry. Charge sharing leads to turning on the Sense Amplifiers followed by activating Cout. Upon charge sharing and sense amplification, the result is Majority(A, B, Cin). \textoverline{Cout} is obtained with the help of a Dual Contact Cell \cite{ambit}. Cin stores the carry result for the next addition computation. The next step is quadruple row activation of  A-1, B-1, Cin, and \textoverline{Cout}, \textoverline{Cout}. After charge sharing and Sense Amplification, P1 is activated to store the result of Majority(A-1, B-1, Cin, \textoverline{Cout}, \textoverline{Cout}). Cin is copied to Cin-1 for storing the same value. For computing the result of the final column, A\textsubscript{1} and B\textsubscript{1} are copied to A and A-1 respectively. Similar to before, the AND-WL is activated followed by turning on Sense Amplifiers. A and A-1 are the activated to store the result of A\textsubscript{1} AND B\textsubscript{1}. row0 is then copied to B and B-1 by activating them. This is done for adding the AND result with the carryin to compute the final sum and carry out. Triple row activation of A, B and Cin and sense amplification is followed by activating P3 and Cout to calculate the final carry out and storing it in P3. The final step is to do a quadruple row activation of A-1, B-1, Cin, \textoverline{Cout}, \textoverline{Cout} and sense amplification followed by activating P2 to store the calculated sum. This sums up the multiplication of 2 bit operands and the final 4 bit results are stored in P0-P3.\newline
\indent It is observed that there are (1+2+3+..+(n-1))*2+n AND operations required where n refers to the number of bits in each operand. On the other hand, there are (1+2+3+...(n-2))*2+n-1+1 total add operations required for an n-bit multiplication operation. In addition to the add operation, we need an initial copy operation for writing 0's to row0. Each copy operation requires 1 AAP. This leads to each AND and add operation requiring 3 AAP operations. Taking all the operations into account, it is found that an n-bit multiplication requires \textbf{3n\textsuperscript{2}+3(n-1)\textsuperscript{2}+4} AAP operations. 
\begin{figure}[htb]
    \centering
    \vspace*{-2pt} 
    \includegraphics[width=0.5\columnwidth]{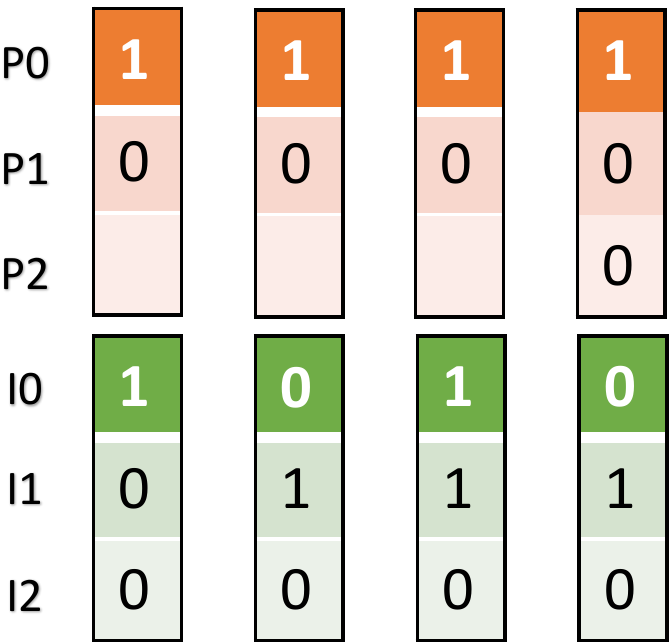}
    \caption{DRAM subarray view}
	\vspace*{-2pt}
    \label{fig:newadd}
\end{figure}
\\\indent This equation holds true when the value of n is lesser than equal to 2. For values of n greater than 2, the addition needs to be performed differently. This is because while adding the AND results of a single column of an n bit number for n greater than 2, the carry can become multiple bits. This changes the multiplication primitive to some extent because the ADD is performed in a different way. Let us look at computing P2 from Fig \mbox{\ref{fig:mulex}}. We need additional n-1 rows for storing the intermediate ADD results. Since we are looking at the example of a 4-bit number, the number of rows required for storing the intermediate ADD results is 3. Let the 3 rows be denoted by I0-I2. Let us imagine that the two operands are 1111 and 1111 in binary. The added result for the column of P1 is 010. 0 is stored in P1 and 001 gets stored in I0-I2. For the result of the column for P2, first the AND of A2 and B0 is computed. The AND results are stored in A and A-1. The previous sum results are copied from I0-I2. An n-1 bit addition \mbox{\cite{mustafa}} is performed and the result is stored back in I0-I2.  The two operands for the ADD operation are the AND result and the previously stored value in I0-I2. The LSB of the first operand used for addition is the AND result. The rest of the bits are considered 0 and are copied from row0 during computation of ADD. The result is stored back in I0-I2. This takes 4*(n-1) AAP operations. This is followed by performing the AND of A1 and B1. Subsequently, we perform the ADD of A1B1 and the previously stored value in I0-I2. The new computed value is stored back in I0-12. Finally, the AND of A0 and B2 is computed. The final ADD of A0B2 and the previously stored value in I0-I2 is computed. The LSB of the final added result is stored in P2. The rest of the bits are stored in I0-I2 for the computation of P3. It is observed that each ADD operation takes 4*(n-1) AAP for an n bit multiplication. Taking this into account, the  total number of AAP operations for n greater than 2 is \textbf{3n\textsuperscript{2}+4(n-1)\textsuperscript{3}+4(n-1)}. Fig \mbox{\ref{fig:newadd}} describes the above mentioned steps. Only P0-P2 instead of P0-P7 is shown in the figure to make the figure more visible.

%% file: sections/exptsetup.tex
We next present the overall PIM-DRAM system design incorporating the hardware architecture and workload data mapping. This section is divided into two subsections. The first subsection describes the architecture of PIM-DRAM. The second subsection explains mapping a neural network model to the PIM-DRAM architecture and the associated dataflow. 

\subsection{PIM-DRAM Architecture}
\begin{figure}[htb]
    \centering
    \vspace*{-2pt} 
    \includegraphics[width=0.6\columnwidth]{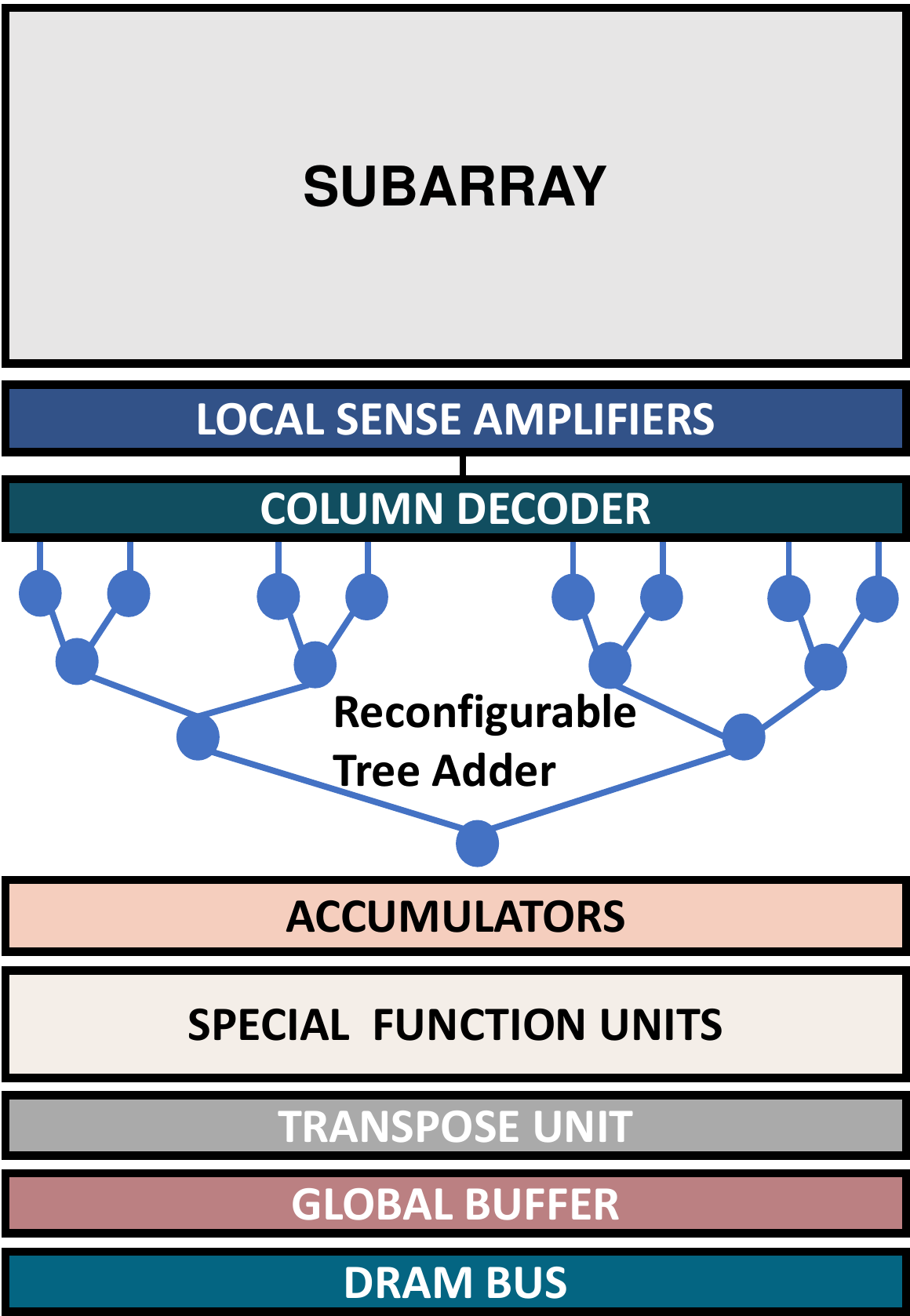}
    \caption{PIM-DRAM bank architecture.}
	\vspace*{-2pt}
    \label{fig:arch}
\end{figure}
PIM-DRAM is comprised of multiple banks connected together through the DRAM internal bus, similar to a conventional DRAM organization. The bank architecture , shown in Fig \ref{fig:arch}, consists of multiple subarrays followed by local sense amplifiers. The proposed multiply operations happen in the subarrays with the data being in transposed format where each column holds the output of an n-bit multiplication. Following multiplication, an accumulation operation is needed to complete the MAC operation. Since the multiplication outputs are in different columns and do not share bitlines, we adopt a reconfigurable adder tree in the peripheral area for intra-bank accumulation. Typically, in our proposed architecture, the sense amplifiers are connected to the reconfigurable adder tree, shown in Fig. \ref{fig:treeadder}, through the column decoder. The adder tree provides  connections from all levels to accumulators for accumulating the sum results. The accumulators are further connected to Special Function Units (SFUs) used for performing non-MVM operations. SFUs include ReLU units, batch-normalization (BatchNorm) units, quantize units and pooling units. Additionally, SFUs are connected to a transposing unit for converting data layout from column-based to row-based and vice versa. These transposing units are connected to the DRAM bus through the global buffer of the bank. The key components of the PIM-DRAM architecture  are described below.

\subsubsection{Reconfigurable Adder Tree}
The reconfigurable adder tree provides both addition and forwarding functionality at each node. Specifically, each node in the adder tree can either forward one operand to the next level node, or add the two operands and pass the result to the next level node.
\begin{figure}[htb]
    \centering
    \vspace*{-2pt} 
    \includegraphics[width=0.7\columnwidth]{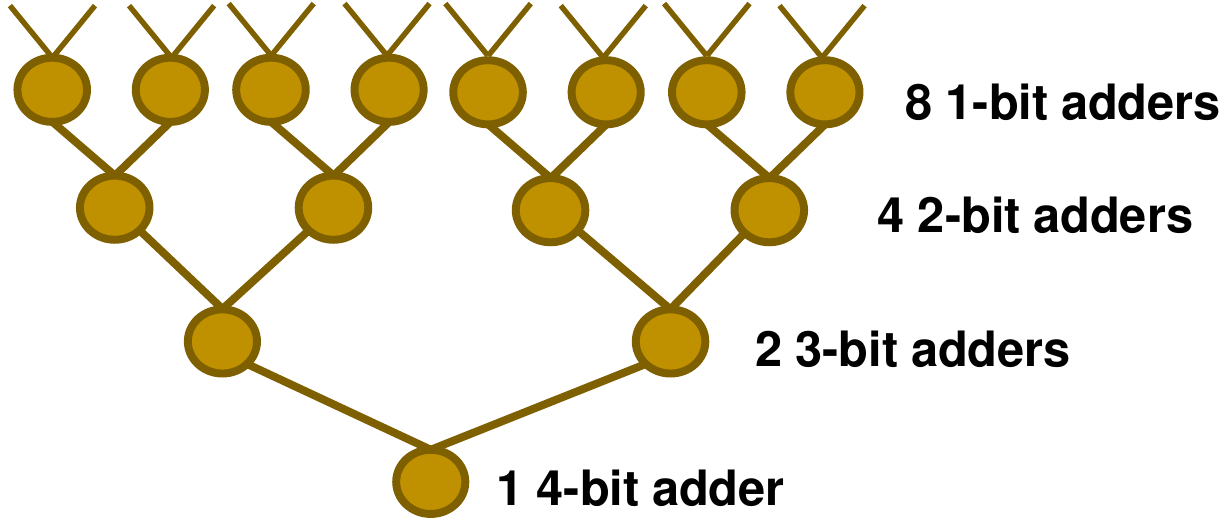}
    \caption{Re-configurable Adder}
	\vspace*{-2pt}
    \label{fig:treeadder}
\end{figure}
The adder tree is connected to the row buffer. The row buffer here is equal to the size of the first level of the adder tree. Each level of the adder tree has a number of units that is a power of 2. The first level has $2$\textsuperscript{n} addition units and the power keeps on decreasing by 1 every level. The last level of the adder has only 1 unit. The example in Fig \ref{fig:treeadder} shows that there are 8 ($2$\textsuperscript{3}) adder units in the first level followed by 4 ($2$\textsuperscript{2}), 2 ($2$\textsuperscript{1}) and 1 ($2$\textsuperscript{0}) in the subsequent levels. Each unit has two inputs and one output. The inputs to the addition units are connected to two outputs of the previous layer. The bit width of the input increases going down the levels of the adder tree. The inputs to the first level of the adder tree are re-configurable. The adder adds the 0\textsuperscript{th} bit till the 2n\textsuperscript{th} bit resulting from an n-bit multiplication. Different Neural Networks have different layer dimensions. A neural network has many different layers which can be convolutional as well as fully connected. Therefore, the number of elements that needs to be added for a single Multiply and Accumulate varies with layers as well as different neural architecture. Therefore, it is essential to have a re-configurable adder to support variable MAC size.
%Adding all the bits are required for computing the final MAC value.
\subsubsection{Accumulators}
Accumulators are units used for accumulating the results of the adder tree. Each accumulator shifts and adds the input received from the adder tree to the existing stored value. For example, when the results from adding the first bit of the MAC arrives, it left-shifts the result by 1 and adds it to the existing stored value from the 0\textsuperscript{th} bit result. The amount to be shifted increases by 1 for every higher order bit and is controlled by a counter. The accumulator accumulates the value till the 2n\textsuperscript{th} addition result arrives. The final result of the MAC is forwarded from the Accumulators to the ReLU units. 
\subsubsection{ReLU Units}
ReLU Activation zeros out any negative sum results and keeps the positive sum values as the same. The output from the accumulators serves as the input to the ReLU Unit. The output data of the ReLU Unit goes to the Batchnorm Unit. 
\subsubsection{Batchnorm Unit}
Batch normalization units take in the inputs from the ReLU Units and standardize the input activations to a layer. The output from this unit is fed to the quantization Unit. Since the batch normalization parameters are fixed for inference, it is a very simple function consisting of subtracting, dividing and scaling by constant factors.
% \subsubsection{Quantize Unit}
% The inputs to this unit come from the Batchnorm units. The Quantize unit quantizes the output to the required precision for the next layer.
\subsubsection{Pooling Units}
Pooling units are only applicable for convolutional layers. For layers that do not have pooling after them, this serves as a pass through unit. Pooling units have a counter that keeps track of how many elements needs to be accounted for the pooling window. It stores a maximum and checks the maximum between the stored maximum and the incoming data and updates the stored maximum. The Pooling unit sits in between the Quantize unit and the Transpose Unit for convolutional layers.
\subsubsection{Transpose Units}
The quantized data needs to converted into the transposed format before being sent out to a different bank to fit the proposed mapping described in the next subsection. The Transpose unit is a 2D array of SRAM cells with dual read and write ports. Data is written horizontally and read out vertically from this unit for transposing it. An example SRAM area for 256x8 is 30534.894 um\textsuperscript{2}.
\begin{table}[h] 
  \begin{center}
    \caption{Area Breakdown}
    \label{table:Area}
    \begin{tabular}{|l|S|S|}
    \hline
      \textbf{Component} & \textbf{Area(um\textsuperscript{2})}&\textbf{Relative Percentage}\\
      \hline
      4096 Adder & 514877 & 99.47373\\
    \hline
    Accumulator & 804 & 0.15532\\
    \hline
    Relu & 431 & 0.083269 \\
    \hline
    Maxpool & 983 & 0.189915 \\
    \hline
    Batchnorm & 506 & 0.097759 \\
    \hline
    Quantize & 91 & 0.017581\\
  
     \hline
    \end{tabular}
 \end{center}
\end{table}
\begin{table}[h] 
  \begin{center}
    \caption{Power Breakdown}
    \label{table:Power}
    \begin{tabular}{|l|S|S|}
    \hline
      \textbf{Component} & \textbf{Power(nW)}&\textbf{Relative Percentage}\\
      \hline
      4096 Adder & 13200190.9 & 95.9014\\
    \hline
    Accumulator & 177765.864 & 1.2915\\
    \hline
    Relu & 109913.671 & 0.7985 \\
    \hline
    Maxpool & 127562.373 & 0.9268 \\
    \hline
    Batchnorm & 120541.29 & 0.8758 \\
    \hline
    Quantize & 28366.738 & 0.2061\\
  
     \hline
    \end{tabular}
 \end{center}
\end{table}

The Area and the Power breakdown of the different components are shown in Table \ref{table:Area} and Table \ref{table:Power}.
\subsection{Data Orchestration}

This section discusses the mapping of a neural network to the PIM-DRAM architecture as well as the associated dataflow. The mapping algorithm is described in Algorithm \ref{alg:Map}.

In the proposed data mapping, every layer is allocated to a DRAM bank. The number of banks required are equal to the number of layers in the network denoted as Number\_of\_Layers. In the mapping algorithm, the outermost loop runs across all the layers in the neural network. Before mapping every layer to a bank, it initializes a MAC\_no (MAC number) and col\_no (column number) variable to 1. For convolution layers, it runs a loop across the number of output filters (no\_output\_filter). For each output filter, there are lot of 3D convolutions associated with it. Each convolution is a MAC operation and each MAC operation involves a significant number of multiplications. The algorithm consists of a nested loop where the outer loop runs across the number of possible convolutions for every filter. That number is obtained by knowing the dimension of the the input width and height, kernel width and height, padding as well as the stride length. The number of iterations (No\_of\_MAC) of the outer loop is given by ((H-K+2*p)/s+1)*(W-L+2*p)/s+1)). Here H and W refer to the input height and width, K and L refer to the kernel height and width, p refers to padding and s refer to stride length respectively. The inner loop runs across the multiply and accumulate operations in a single 3D convolution. Here the number of iterations (MAC\_size) is given by K*L*I where I is the number of input channels. The mapping is done by assigning the same MAC number (MAC\_no) to the operands in the same multiply and accumulate.

Every multiplication of a MAC is mapped to a single column and the column number (col\_no) is increased for the mapping the next multiplication. After mapping all the multiplications of the same multiply and accumulate, the MAC\_no is increased by 1. The mapping starts with the first subarray. The subarray count (sub\_no) is increased by 1 and the column\_no is set to 1 when the column limit is reached for the particular subarray. 
\begin{algorithm}
\caption{Mapping algorithm for DNN's}
\begin{algorithmic} 
\REQUIRE Network Description
\WHILE{$layer\_no \leq Number\_of\_Layers$}
\IF{$layer[layer\_no] == conv$}
\FOR{$i=1; i \leq no\_output\_filter; i=i+1$}
\IF{$i\%(no\_output\_filter \div K)==0$}
\STATE $sub\_no \leftarrow 1,col\_no \leftarrow 1$
\ENDIF
\FOR{$j=1;j \leq No\_of\_MAC;j=j+1$}
\IF{$col\_no + MAC\_size \leq column\_size$}
\FOR{$k=1;k \leq MAC\_size;k=k+1$}
\STATE$Bank[sub\_no][col\_no] \leftarrow MAC\_no$
\STATE $col\_no \leftarrow col\_no+1$

\ENDFOR

\ENDIF
\IF{$col\_no + MAC\_size > column\_size$}
\STATE $sub\_no \leftarrow sub\_no+1,col\_no \leftarrow 1$
\FOR{$k=1;k \leq MAC\_size;k=k+1$}
\STATE$Bank[sub\_no][col\_no] \leftarrow MAC\_no$
\STATE $col\_no \leftarrow col\_no+1$

\ENDFOR
\ENDIF
\STATE $MAC\_no \leftarrow MAC\_no+1$

\ENDFOR
\ENDFOR
\ENDIF
\IF{$layer[layer\_no] == linear$}
\FOR{$i=1;i< no\_output\_neuron;i=i+1$}
\IF{$i\%(no\_output\_neuron \div k)==0$}
\STATE $sub\_no \leftarrow 1,col\_no \leftarrow 1$ 
\ENDIF
\IF{$col\_no + MAC\_size \leq column\_size$}
\FOR{$j=1; j \leq MAC\_count; j=j+1$}
\STATE$Bank[sub\_no][col\_no] \leftarrow MAC\_no$
\STATE $col\_no \leftarrow col\_no+1$
\ENDFOR
\ENDIF
\IF{$col\_no + MAC\_size > column\_size$}
\STATE $sub\_no \leftarrow sub\_no+1,col\_no \leftarrow 1$
\FOR{$k=1;k \leq MAC\_size;k=k+1$}
\STATE$Bank[sub\_no][col\_no] \leftarrow MAC\_no$
\STATE $col\_no \leftarrow col\_no+1$

\ENDFOR
\ENDIF
\STATE $MAC\_no \leftarrow MAC\_no+1$
\ENDFOR
\ENDIF
\STATE $layer\_no \leftarrow layer\_no + 1$
\ENDWHILE
\end{algorithmic}
 \label{alg:Map}
\end{algorithm}

One of the rules of the mapping algorithm is that all the operands under the same MAC must be in the same subarray. This is done to fit all the operands of the MAC in the same adder tree. While mapping the operands to the columns of the subarray, if the number of multiplications in the MAC is greater than the remaining columns in the subarray, then the mapping starts from column 1 in the next subarray and nothing gets mapped to the remaining columns in the previous subarray. This rule is applicable for mapping linear layers as well. Every pair of operands in a mapped column in the subarray has an n bit activation and a corresponding n bit weight value occupying 2n rows altogether.

\begin{figure*}[htb]
  \vspace*{-0pt}
  \centering
  \includegraphics[width=0.9\textwidth]{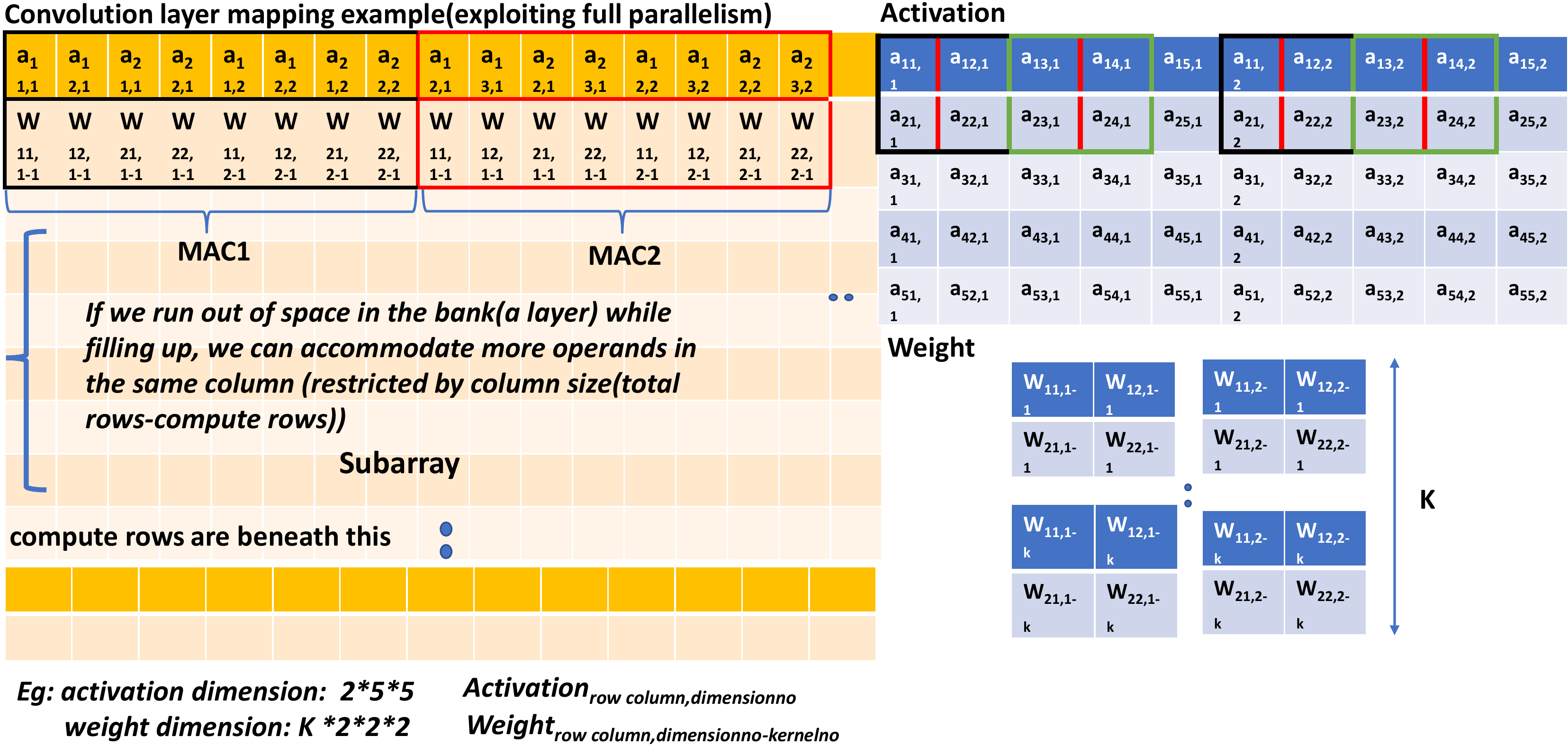}
  \vspace*{-2pt}
  \caption{Mapping example of convolution layer}
  \label{fig:Mapeg}
  \vspace*{-8pt}
\end{figure*}

Having just one pair of operands in all the columns provides the maximum amount of parallelism. It is also possible that the mapping may run of space due to limited capacity of a bank. In that case, the mapper can divide the number of output filters into k groups where the number of output filters is divisible by k. In that case after every (no\_output\_filter)/k filters, it goes back to subarray 1 and column 1 and starts mapping from there again. In this way, it puts more pairs of operands in a single column which needs to be processed sequentially. The amount of parallelism reduces as a result of that. Higher the value of k, the lesser is the parallelism. An example mapping of a convolution layer is shown in Fig \ref{fig:Mapeg}. The mapping of a fully connected layer is very similar, with some minute differences. The outer loop in the fully connected layer runs across all the output neurons (no\_output\_neuron).

The mapping algorithm divides the number of output neurons into k groups. After every (no\_output\_neuron/k) neurons, it goes back to subarray 1 and column 1 for mapping the next set of neurons. 
The inner loop runs across the connections to a single output neuron. Similar to the convolution layer case, the same MAC number is given to all the multiplications in a particular multiply and accumulate. One more difference with regards to convolution layer mapping is that the parallelism is controlled by the number of output neurons instead of the number of output kernels. Higher parallelism comes with the cost of bad memory footprint. The worst case memory footprint for a convolutional layer is given by O*((H-K+2*p)/s+1) *((W-L+2*p)/s+1)*(I*L*K)*2*n where n is the number of bits of each operand and O is the number of output filters. We can get lesser footprint by reusing some operands and stacking up more operands in the same columns. But that comes with the cost of lesser parallelism. So, there is a trade of between parallelism and memory footprint.The worst case memory footprint of a fully connected layer is w1*w2*2*n where w1*w2 are the weight dimensions of a fully connected layer.
\\ \indent Next we briefly discuss the dataflow for the above mentioned mapping and the proposed architecture. The dataflow follow a fixed order for every bank. Different banks work in a pipelined manner. As every bank is associated with a different layer of the neural network, they all can work on different images in the dataset at the same time. Taking an example of a 3 layer network, bank 3 will be working on the k\textsuperscript{th} image when bank 2 and bank 1 will be working on the k-1\textsuperscript{th} and k-2\textsuperscript{th} image respectively. The initial operation in a bank begins with the multiply operation happening across all subarrays in every bank. The bits of the product in every column is read by the the adder tree. The number of MACs of a subarray to be added depends on how many MACs can be fit in the adder. The adder tree keeps on adding results of the products from 0\textsuperscript{th} till the 2n\textsuperscript{th} bit. The accumulators work in conjunction with the adder accumulating all the results of the adder. Once the accumulator has accumulated all the results, the final MAC values are sent to the special function units where each logic block takes its required amount of cycles. Following that, the data is fed into the transpose units. All the banks work in parallel until this point. Then the banks transfer data sequentially using Rowclone \cite{rowclone} to the destination banks. Again, taking the example of a 3 layer network, bank 2 will send its data to bank 3 followed by bank 1 sending its data to bank 2, and so on. After the sequential data transfers, the adder works on the remaining MACs and the same set of operations except the multiplication operation are performed again. In cases where there are more than one pair of operands in a single column, they get executed sequentially over time.
\begin{figure}[htb]
    \centering
    \vspace*{-2pt} 
    \includegraphics[width=\columnwidth]{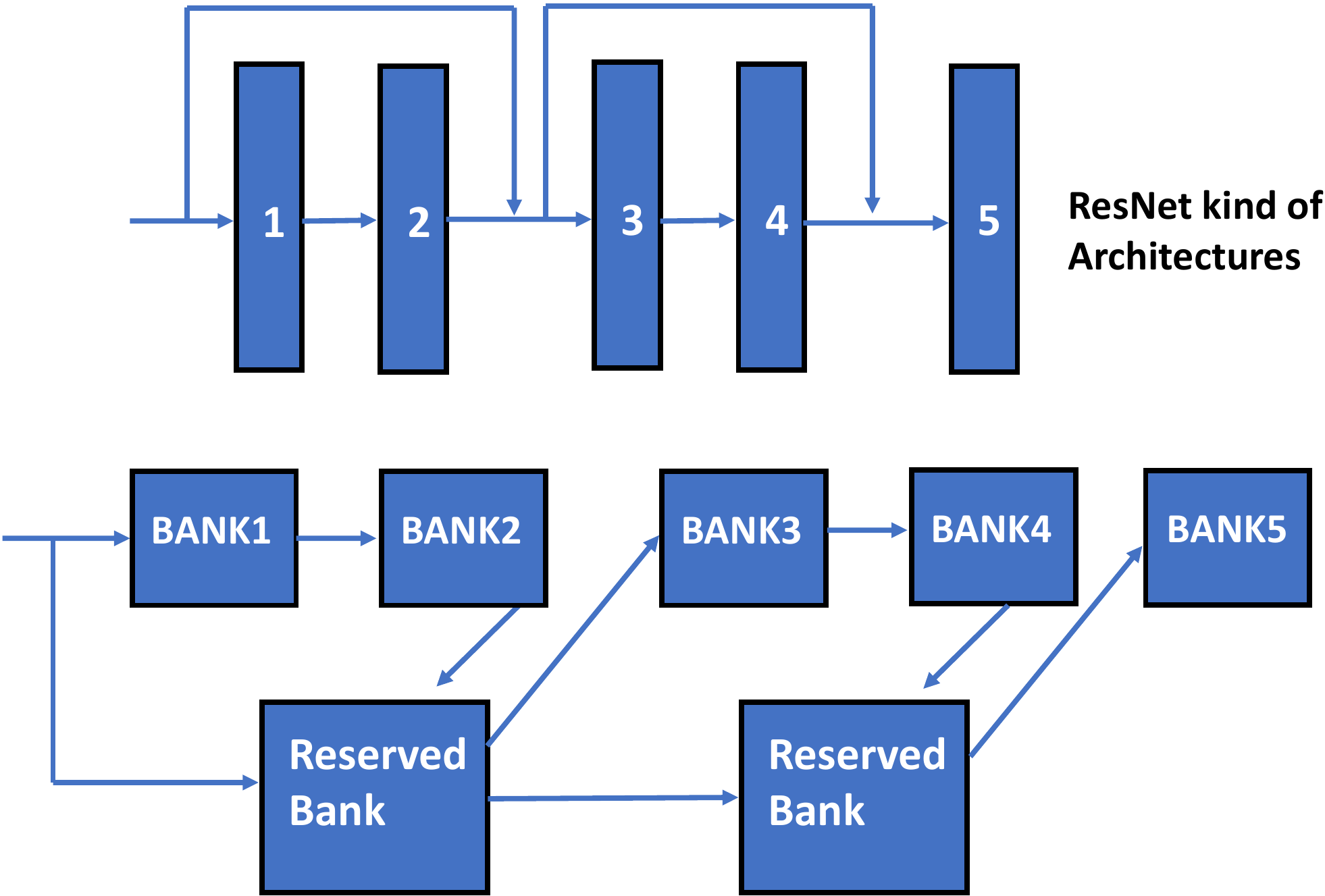}
    \caption{ResNet Mapping}
	\vspace*{-2pt}
    \label{fig:resnet}
\end{figure}
Residual layers reserve some banks for adding the results of the skip connections with the output activations of a layer to calculate the final output activations. For residual connections, the shortcut inputs are copied to a Reserved Bank using Rowclone \mbox{\cite{rowclone}}. Then after the computing activation from different layers, the output is copied to the same Reserved Bank. The operands are added using \mbox{\cite{mustafa}} and transferred to the respective destination bank as shown in Fig \mbox{\ref{fig:resnet}}. 

%% file: sections/results.tex
\subsection{Circuit-Level Analysis}
We perform extensive circuit simulations using HSPICE to evaluate the proposed in-DRAM compute primitives. Our simulations are performed in CMOS 65 nm technology with the DRAM cells and sense amplifier parameters adapted from the Rambus power model \cite{rambus}. 

\begin{figure}[htb]
    \centering
    \vspace*{-2pt} 
    \includegraphics[width=0.9\columnwidth]{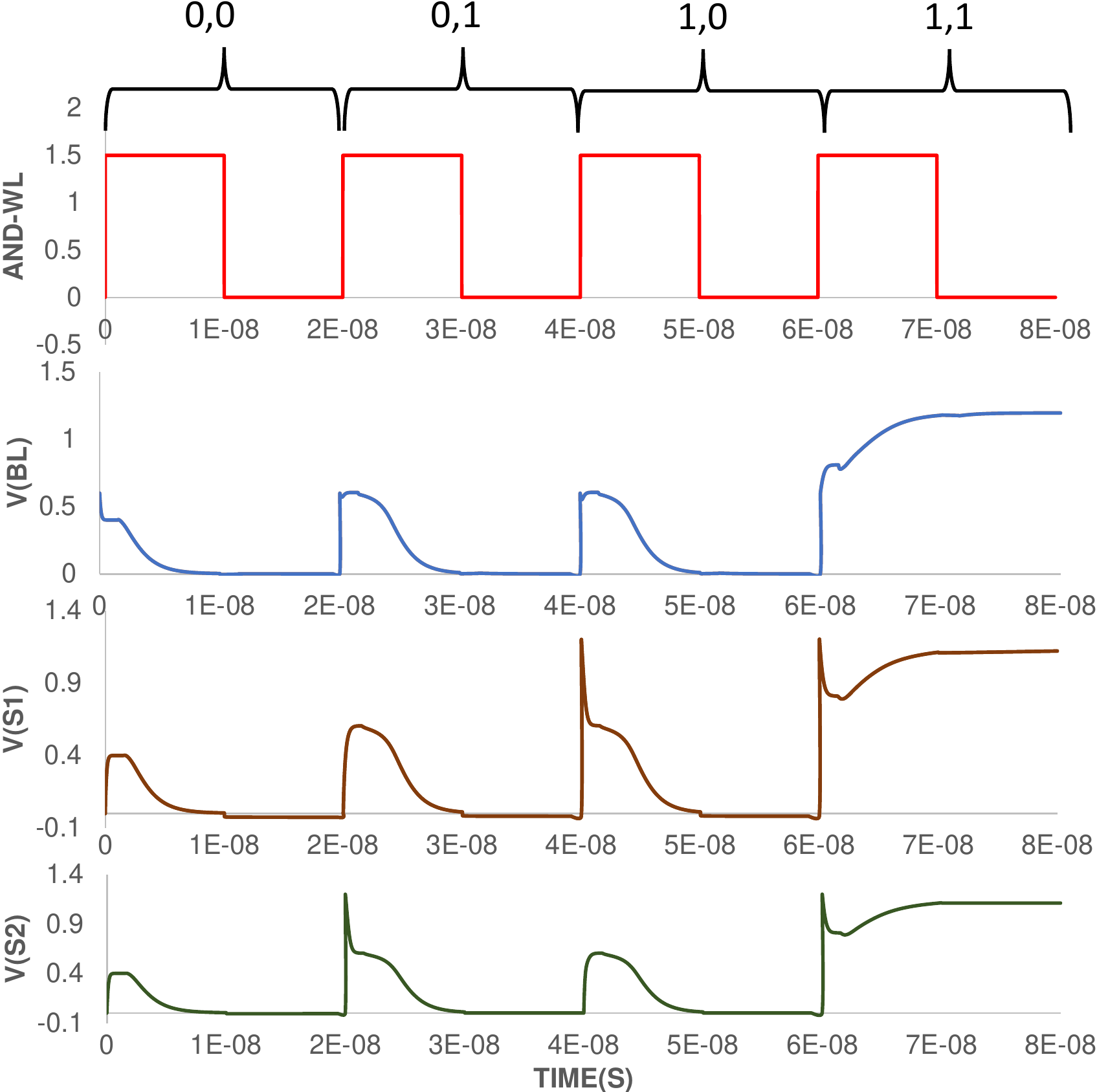}
    \caption{Spice Simulations}
	\vspace*{-2pt}
    \label{fig:spice}
\end{figure}
Fig. \ref{fig:spice} shows the transient analysis of the proposed AND operation in DRAM subarrays for all input combinations. S1, and S2 are the nodes of the top plates of the two cell capacitors. For the {1,1} case BL, S1, and S2 nodes reach VDD, while in other cases the corresponding nodes drop to GND, representing the AND operation.

The waveforms shown in Fig. \ref{fig:spice} 
\begin{figure}[htb]
    \centering
    \vspace*{-2pt} 
    \includegraphics[width=\columnwidth]{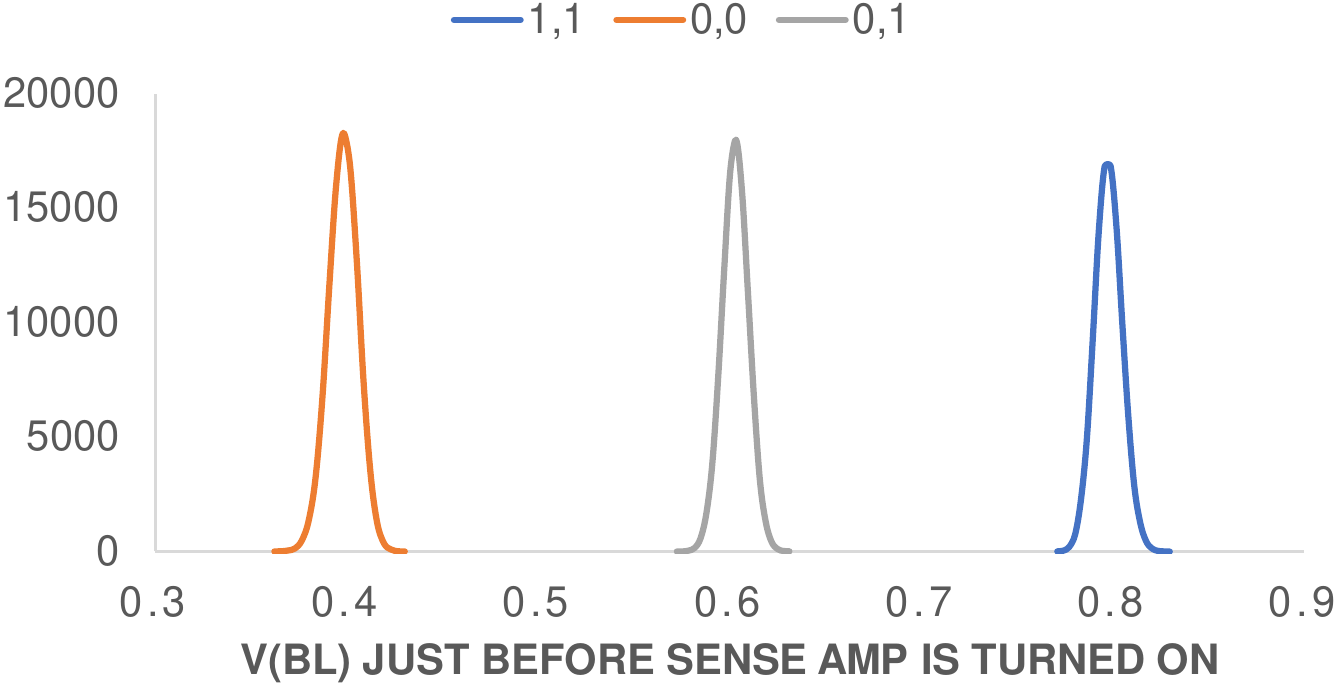}
    \caption{Monte Carlo for 100000 samples}
	\vspace*{-2pt}
    \label{fig:monte}
\end{figure}

Furthermore, we perform 100000 Monte Carlo simulations of the AND operation with all input cases to validate the robustness of our proposed PIM primitive. Fig. \ref{fig:monte} shows the histograms of all input cases of the BL node before enabling the sense amplifier. We observe large enough sense margin of BL between all input cases (mean is 200mV). 
%What variations did you assume? In the access transsistor, capacitor? What sigma-to-mu?

\subsection{System-level Analysis}
We developed our in-house simulator to evaluate the proposed PIM DRAM architecture running commonly-used machine learning workloads (AlexNet, VGG-16, and ResNet-18). We consider DDR3-1600 DRAM structure in our system analysis with a subarray size of 4096x4096. Our simulator maps the workload layers to the DRAM based on layer size to optimize performance. Moreover, it considers all performed operations in the DRAM including computing and internal data movements. We compare the proposed PIM DRAM with GPU.
\begin{comment}
The ideal non-PIM system has a compute unit capable of instantaneous computation, coupled with a conventional DRAM, so that the only latency cost is coming from moving the data between the host and CPU. Note that the ideal non-PIM baseline represents the upper bound on performance for any non-PIM system (e.g., GPU and TPU).
\end{comment}
We model all the additional logic blocks using RTL and synthesize them using the Cadence RTL Compiler to the TSMC 65 nm library. To consider the effect of DRAM process on logic blocks performance, we add 21.5\% delay to each block based on the study reported in \cite{logic}.

\begin{figure}[htb]
    \centering
    \vspace*{-2pt} 
    \includegraphics[width=\columnwidth]{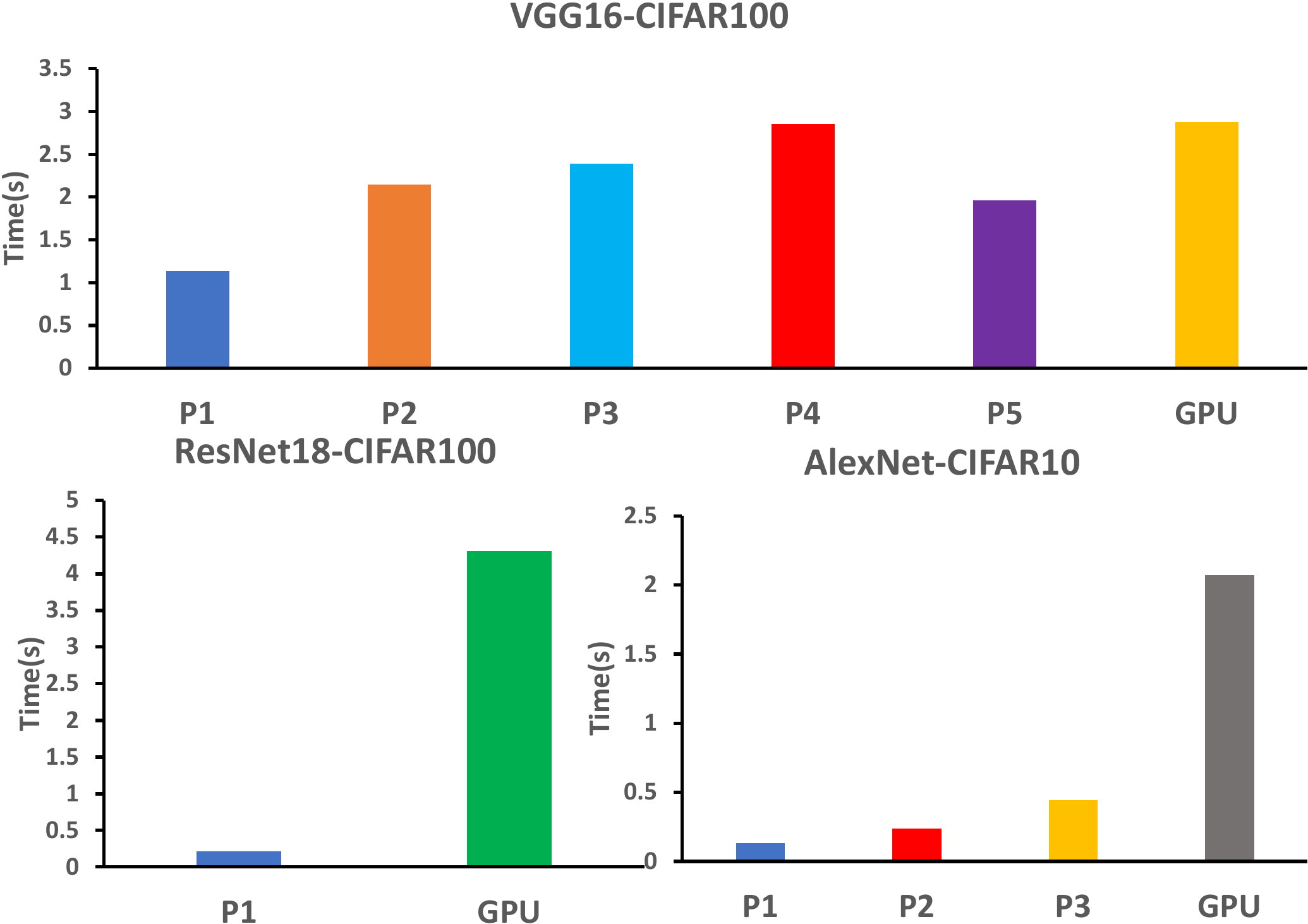}
    \caption{System Level Evaluation}
	\vspace*{-2pt}
    \label{fig:sys}
\end{figure}
\begin{figure}[htb]
    \centering
    \vspace*{-2pt} 
    \includegraphics[width=\columnwidth]{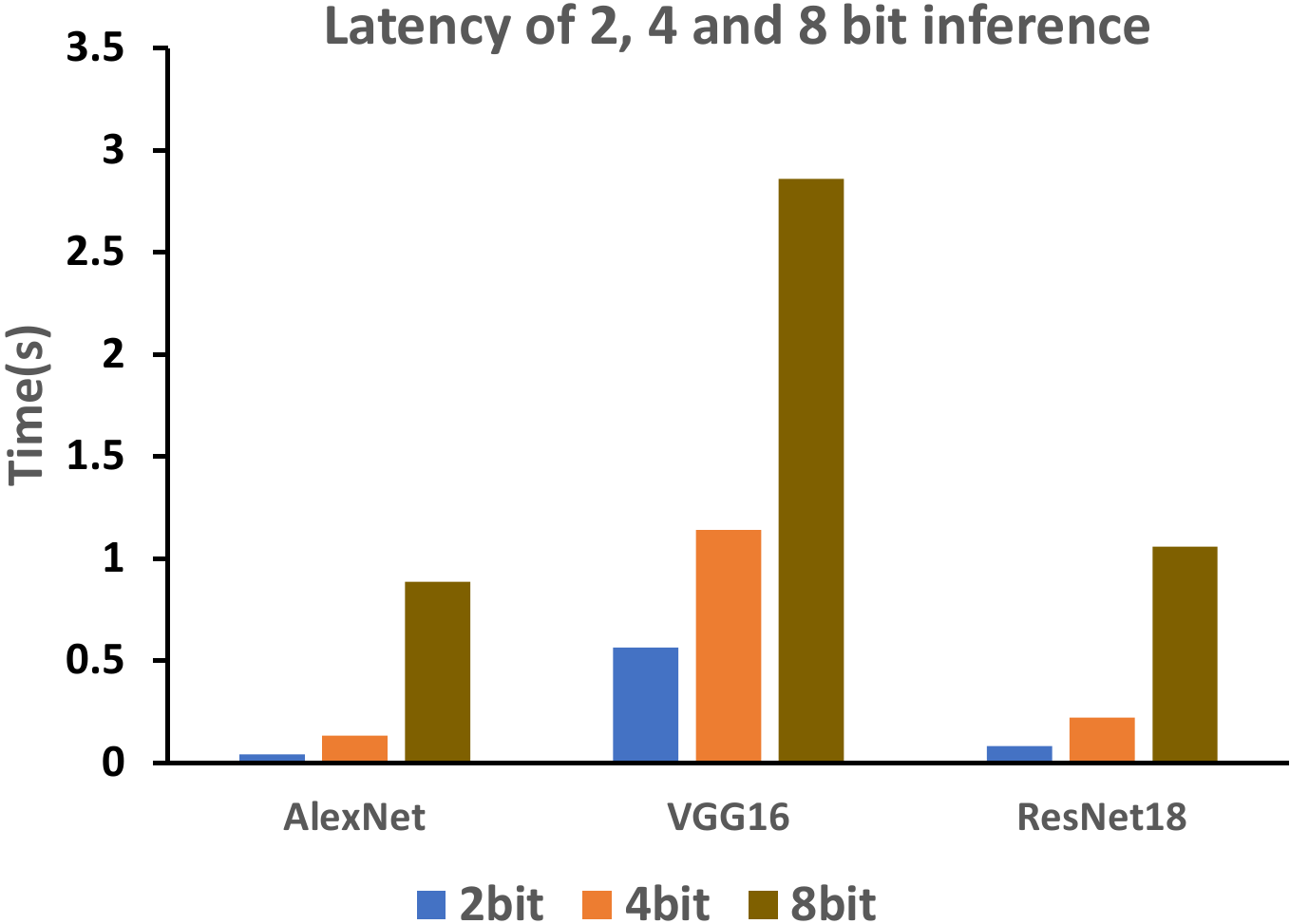}
    \caption{Simulation time for different bit precision}
	\vspace*{-2pt}
    \label{fig:scale}
\end{figure}
Fig \ref{fig:sys} shows the performance benefits of our proposed PIM architecture over GPU. The GPU we used in our evaluation is NVIDIA TITAN Xp. It has 3840 CUDA Cores, a Memory Speed of 11.4 Gbps and a Memory Bandwidth of 547.7 GB/s. Moreover, we vary the parallelism of our PIM architecture in our comparisons. P1, P2, P3 and P4 refer to the parallelism factor for each layer of the the neural network described in Section \ref{sec:exptsetup} . For AlexNet, P1 refers to (1, 1, 1, 1, 1, 1, 1, 1), P2 refers to (2, 2, 2, 2, 2, 2, 2, 2) and P3 refers to (4, 4, 4, 4, 4, 4, 2, 1). Here, the 8 numbers refer to the parallelism factors for the 8 layers of AlexNet. For VGG16 P1, P2, P3, P4 and P5 refer to (1, 1, 1, 1, 1, 1, 1, 1, 1, 1, 1, 1, 1, 1, 1, 1), (2, 2, 2, 2, 2, 2, 2, 2, 2, 2, 2, 2, 2, 2, 2, 2), (4, 4, 4, 4, 4, 4, 4, 4, 4, 4, 4, 4, 4, 4, 4, 4), (8, 8, 8, 8, 8, 8, 8, 8, 8, 8, 8, 8, 8, 4, 4, 4) and (8, 8, 8, 8, 8, 8, 8, 8, 8, 8, 8, 8, 8, 1, 1, 1) respectively for the 16 layers. Finally for ResNet18, P1 refers to (1, 1, 1, 1, 1, 1, 1, 1, 1, 1, 1, 1, 1, 1, 1, 1, 1, 1), for the 18 layers. We have considered 4-bit weights and activations in our evaluation. Fig \mbox{\ref{fig:scale}} shows how the performance scales with different bit precision.

The proposed PIM-DRAM shows performance benefits over GPU on all networks. We achieve up to 19.5x peak speedup over ideal GPU, respectively.

%% file: sections/conclusion.tex
{\noindent} 
Processing-in-memory is a promising solution to the well-known memory wall problem in current machine learning hardware. We propose a new in-subarray PIM-DRAM computing architecture specifically-tailored for machine learning workloads. Since MAC operations consume the dominant part of most ML workload runtime, we propose in-subarray multiplication coupled with intra-bank accumulation. The multiplication operation is performed by performing AND operations and addition in column-based fashion while only adding less than 1\% area overhead. Moreover, the proposed multiplication primitive can be easily adopted in commodity DRAM chips since it requires no modifications to the DRAM peripheral circuitry. We incorporate a reconfigurable adder tree to perform intra-bank accumulation. Moreover, special function units are incorporated in each bank to perform the element-wise operations required in ML workloads (e.g. batchnorm, ReLU, \textit{etc}). We propose an efficient data mapping algorithm to schedule the MVM operations on corresponding DRAM banks and subarrays. The proposed DRAM-based PIM archietcure and dataflow are evaluated on common ML workloads and provide 19.5x better performance compared to a GPU, respectively.
%% non-PIM baseline??
% Processing in memory is a promising solution to address the memory bound problem of data intensive applications on edge devices which are energy and storage constrained. In this work, we proposed an in-memory computing primitive for AND and MULTIPLICATION operations. Our work makes use of the primitives to propose a DRAM architecture consisting of re-configurable adder trees, accumulators, ReLU units, batchnorm units, quantize units and transpose units. This work also proposes a suitable mapping technique to map DNN's on to this architecture and a dataflow to maximize parallelism. We validated our primitive using Spice Simulations. We built an in house simulator to evaluate DNN's like AlexNet, VGG16 and ResNet18 on datasets such as CIFAR10 and CIFAR100 . Our evaluations show that PIM-DRAM can provide benefits up to 23x over GPU and 6.5x over Ideal Non Pim.